\documentclass{article} % For LaTeX2e
\usepackage{hyperref,fullpage}
\usepackage{url}

\usepackage{amsmath,amssymb,bm}
\usepackage{graphicx}
\usepackage{rotating}
\usepackage{multirow,subfigure}
\usepackage{appendix}

\newcommand{\ie}{\emph{i.e.}}
\newcommand{\eg}{\emph{e.g.}}
\newcommand{\etal}{\emph{et al.}}

\newcommand{\couldcut}[1]{}
\usepackage{amsmath,amsfonts}
\usepackage{amssymb,amsopn}
\usepackage{bm} % bold symbol

\usepackage{amssymb}
\usepackage{amsmath,amsfonts}
\usepackage{amsthm,amsopn}

\usepackage{bm} % bold symbol

\newcommand{\vct}[1]{\boldsymbol{#1}} % vector
\newcommand{\mat}[1]{\boldsymbol{#1}} % matrix
\newcommand{\cst}[1]{\mathsf{#1}}  % constant

\newcommand{\field}[1]{\mathbb{#1}}
\newcommand{\R}{\field{R}} % real domain
\newcommand{\T}{^{\textrm T}} % transpose

\newcommand{\twonorm}[1]{\left\|#1\right\|_2^2}

\newcommand{\ProbOpr}[1]{\mathbb{#1}}
\newcommand{\expect}[2]{%
\ifthenelse{\equal{#2}{}}{\ProbOpr{E}_{#1}}
{\ifthenelse{\equal{#1}{}}{\ProbOpr{E}\left[#2\right]}{\ProbOpr{E}_{#1}\left[#2\right]}}} % Expectation: syntax: E{1}{2} = E_1[2], E{}{2}=E[2], E{1}{} = E_1
\newcommand{\var}[2]{%
\ifthenelse{\equal{#2}{}}{\ProbOpr{VAR}_{#1}}
{\ifthenelse{\equal{#1}{}}{\ProbOpr{VAR}\left[#2\right]}{\ProbOpr{VAR}_{#1}\left[#2\right]}}} % Expectation: syntax: V{1}{2} = V_1[2], V{}{2}=V[2], V{1}{} = V_1
\DeclareMathOperator{\argmax}{arg\,max}

\newcommand{\Xt}{{\mathcal{X}_{(t)}}} % X_t
\newcommand{\Ystart}{{Y^{\star}_{(t)}}} % Y*_t
\newcommand{\ground}{{\mathcal{Y}}} % ground set
\newcommand{\groundX}{{\mathcal{X}}} % ground set

\newcommand{\vtheta}{\vct{\theta}}

\newcommand{\vx}{{\vct{x}}}
\newcommand{\vy}{\vct{y}}

\newcommand{\cN}{\cst{N}}

\newcommand{\mL}{\mat{L}}
\newcommand{\mI}{\mat{I}}
\newcommand{\mK}{\mat{K}}

\newcommand{\vzero}{\vct{0}}
\newcommand{\vphi}{\vct{\phi}}

\newcommand{\eat}[1]{}

\title{Large-Margin Determinantal Point Processes}

\author{
Boqing Gong\\
Dept.\ of Computer Science\\
U.\ of Southern California\\
Los Angeles, CA 90089\\
\texttt{boqinggo@usc.edu} 	
\and
Wei-lun Chao\\
Dept.\ of Computer Science\\
U.\ of Southern California\\
Los Angeles, CA 90089\\
\texttt{weilunc@usc.edu} 	
\and
Kristen Grauman\\
Dept.\ of Computer Science\\
U.\ of Texas at Austin\\
Austin, TX 78701 \\
\texttt{grauman@cs.utexas.edu}	
\and
Fei Sha\\
Dept.\ of Computer Science\\
U.\ of Southern California\\
Los Angeles, CA 90089\\
\texttt{feisha@usc.edu}  		
}

\makeatletter
\renewcommand{\paragraph}{%
  \@startsection{paragraph}{4}%
  {\z@}{0ex \@plus 1ex \@minus .2ex}{-1em}%
  {\normalfont\normalsize\bfseries}%
}
\makeatother
\begin{document}

\maketitle

%%%%%%%%%%%%%%%%%%%%%%%%%%%%%%%%%%%%%%%%%%%%%%%%%%%%%%%%%%%%%%%%%%%%
%%%%%%%%%%%%%%%%%%%%%%%%%%%%%%%%%%%%%%%%%%%%%%%%%%%%%%%%%%%%%%%%%%%%
\begin{abstract} 
% !TEX root = main.tex

Determinantal point processes (DPPs) offer a powerful approach to modeling diversity in many applications where the goal is to select a diverse subset.  We study the problem of learning the parameters (\ie, the kernel matrix) of a DPP from labeled training data.  We make two contributions.  First, we show how to reparameterize a DPP's kernel matrix with multiple kernel functions, thus enhancing modeling flexibility.  Second, we propose a novel parameter estimation technique based on the principle of large margin separation.   In contrast to the state-of-the-art method of maximum likelihood estimation, our large-margin loss function explicitly models errors in selecting the target subsets, and it can be customized to trade off different types of errors (precision vs.~recall). Extensive empirical studies validate our contributions, including applications on challenging document and video summarization, where flexibility in modeling the kernel matrix and balancing different errors is indispensable.

\end{abstract} 

% !TEX root = main.tex
\section{Introduction}
\label{sIntro}

Imagine we are to design a search engine to retrieve web images that match user queries. In response to the search term \textsc{jaguar}, what should we retrieve --- the images of the animal jaguar or the images of the automobile jaguar?

This frequently cited example illustrates the need to incorporate the notion of \emph{diversity}. In many tasks, we want to select a subset of items from a ``ground set''. While the ground set might contain many similar items, our goal is \emph{not} to discover all of the same ones, but rather to find a subset of diverse items that ensure coverage (the exact definition of coverage is task-specific). In the example of retrieving images for \textsc{jaguar}, we achieve diversity by including both types of images. %, and avoid sets of images containing \emph{only one} type of the two.

Recently, the determinantal point process (DPP) has emerged as a promising technique for modeling diversity~\cite{kulesza2012determinantal}. A DPP defines a probability distribution over the power set of a ground set. Intuitively, subsets of higher diversity are assigned larger probabilities, and thus are more likely to be selected than those with lower diversity. Since its original application to quantum physics, DPP has found many applications in modeling random trees and graphs~\cite{burton1993local},  document summarization~\cite{kulesza2011learning}, search and ranking in information retrieval~\cite{kulesza2011k}, and clustering~\cite{kang2013fast}. Various extensions have also been studied, including k-DPP~\cite{kulesza2011k}, structured DPP~\cite{kulesza2011structured}, Markov DPP~\cite{affandi2012markov}, and DPP on continuous spaces~\cite{affandi2013approximate}. 

The probability distribution of a DPP depends crucially on its kernel --- a square and symmetric, positive semidefinite matrix whose elements specify how similar every pair of items in the ground set are. This kernel matrix is often unknown and needs to be estimated from training data. 

This is a very challenging problem for several reasons.  First, the number of the parameters, i.e., the number of elements in the kernel matrix, is quadratic in the number of items in the ground set. For many tasks (for instance, image search), the ground set can be very large. Thus it is impractical to directly specify every element of the matrix, and a suitable reparameterization of the matrix is necessary. Secondly, the number of training samples is often limited in many practical applications. One such example is the task of document summarization, where our aim is to select a succinct subset of sentences from a long document. There, acquiring accurate annotations from human experts is costly and difficult. Thirdly, for many tasks, we need to evaluate the performance of the learned DPP not only by its accuracy in predicting whether an item should be selected, but also by other measures like precision and recall. For instance, failing to select key sentences for summarizing documents might be regarded as being more catastrophic than injecting sentences with repetitive information into the summary. 

Existing methods of parameter estimation for DPPs are inadequate to address these challenges.  For example, maximum likelihood estimation (MLE) typically requires a large number of training samples in order to estimate the underlying model correctly. This also limits the number of the parameters it can estimate reliably, restricting its use to DPPs whose kernels can be parameterized with few degrees of freedom. It also does not offer fine control over  precision and recall. %\FS{Should we say Affandi's Bayesian approach?}

We propose a two-pronged approach for learning a DPP from labeled data. First, we improve modeling flexibility by reparameterizing the DPP's kernel matrix with multiple base kernels. This representation could easily incorporate domain knowledge and requires learning fewer parameters (instead of the whole kernel matrix). Then, we optimize the parameters such that the probability of the correct subset is larger than other erroneous subsets by a large margin. This margin is task-specific and can be customized to reflect the desired performance measure---for example, to monitor precision and recall. As such, our  approach defines objective functions that closely track selection errors  and work well with few training samples.  While the principle of large margin separation has been widely used in classification~\cite{vapnik} and structured prediction~\cite{taskar}, formulating DPP learning with the large margin principle is novel.  Our empirical studies show that the proposed method attains superior performance on two challenging tasks of practical interest: document and video summarization.

%We show how to leverage the computational properties of DPP to avoid an explicit comparison between the correct subset and an exponential numbers of incorrect subsets. 

The rest of the paper is organized as follows. We provide background on the DPP in section~\ref{sDPP}, followed by our approach in section~\ref{sApproach}.  We discuss related work in section~\ref{sRelated} and report our empirical studies in section~\ref{sExp}. \couldcut{There we demonstrate the advantages of our method on both synthetic and real-world datasets from practical applications.}  We conclude in section~\ref{sConclusion}.

\section{Background: Determinantal point processes}
\label{sDPP}

%\KG{This section needs a ref (or more) early on to attribute the basic model.}
%\BG{1) Added citation to ~\cite{macchi1975coincidence} in the first paragraph. 2) Added ``We refer readers to~\cite{kulesza2012determinantal} for more details.''}
We first review background on the determinantal point process (DPP)~\cite{macchi1975coincidence} and the standard maximum likelihood estimation technique for learning DPP parameters from data. More details can be found in the excellent tutorial~\cite{kulesza2012determinantal}.

Given a ground set of $\cN$ items, $\ground=\{1, 2,\ldots,\cN\}$, a DPP defines a probabilistic measure over the power set, i.e., all possible subsets (including the empty set) of $\ground$. Concretely, let $\mL$ denote a symmetric and positive semidefinite matrix in $\R^{\cN \times\cN}$. The probability of selecting a subset $\vy \subseteq \ground$ is given by
\begin{equation}
P( \vy ; \mL) = \det(\mL+\mI)^{-1}\det(\mL_{\vy}),
\label{eDPPLensemble}
\end{equation}
where $\mL_{\vy}$ denotes the submatrix of $\mL$, with rows and columns selected by the indices in $\vy$. $\mI$ is the identity matrix with the  proper size.  We define $\det{(\mL_\emptyset)}=1$.  The above way of defining a DPP is called an L-ensemble. An equivalent way of defining a DPP is to use a kernel matrix to define the marginal probability of selecting a random subset:
\begin{equation}
P_{\vy} = \sum_{\vy' \subseteq \ground} P(\vy'; \mL) \mathbb{I}[\, \vy \subseteq \vy'\, ] = \det (\mK_{\vy}),\label{ePy}
\end{equation}
where we sum over all subsets $\vy'$ that contain $\vy$ ($\mathbb{I}[\,\cdot\,]$  is an indicator function). 
The matrix $\mK$ is another positive semidefinite matrix, computable from  the $\mL$ matrix 
\begin{equation}
\mK = \mL(\mL + \mI)^{-1},
\label{eDPPKernel}
\end{equation}
and $\mK_{\vy}$ is the submatrix of $\mK$ indexed by $\vy$. Despite the exponential number of summands in eq.~\eqref{ePy}, the marginalization is analytically tractable and computable in polynomial time.  

% Each element of $\mK$ (and $\mL$) specifies how similar the corresponding pair of items in $\ground$ are.
%Note that the maximum  eigenvalue of $\mK$ must be at most 1 such that the determinants of all its principal minors are at most 1, in order for $\det (\mK_{\vy})$ to be a valid probability.

\paragraph{Modeling diversity} One particularly useful property of the DPP is its ability to model \emph{pairwise repulsion}. Consider the marginal probability of having two items $i$ and $j$ simultaneously in a subset:
\begin{equation}
P_{\{i,j\}} = \det\left|\begin{aligned}
K_{ii} & K_{ij}\\
K_{ji} & K_{jj}\end{aligned}\right| = K_{ii}K_{jj} - K_{ij}^2 \le K_{ii}K_{jj} = P_{\{i\}}P_{\{j\}} \le \min( P_{\{i\}}, P_{\{j\}}).
\end{equation} 
Thus, unless $K_{ij}=0$, the probability of observing $i$ and $j$ jointly is always less than observing either $i$ or $j$ separately.   Namely, having $i$ in a subset repulsively excludes $j$ and vice versa.  Another  extreme case is when $i$ and $j$ are the same; then $K_{ii} = K_{jj} = K_{ij}$, which leads to $P_{\{i,j\}} = 0$. Namely, we should never allow them together in any subset.

Consequently, a subset with a large (marginal) probability cannot have too many items that are similar to each other (i.e., with high values of $K_{ij}$). In other words, the probability provides a gauge of the diversity of the subset. The most diverse subset, which balances all the pairwise repulsions, is the subset that attains the highest probability 
\begin{equation}
\vy^* = \argmax_{\vy} P(\vy; \mL).
\label{eDPPMAP}
\end{equation}
Note that this MAP inference is computed with respect to the L-ensemble (instead of $\mK$) as we are interested in the mode, not the marginal probability of having the subset.  Unfortunately, the MAP inference is NP-hard~\cite{ko1995exact}.  Various approximation algorithms have been investigated~\cite{gillenwater2012near,kulesza2012determinantal}.

\paragraph{Maximum likelihood estimation (MLE)} %How to determine the $\mL$ or $\mK$ matrices for a ground set? Appealing to domain knowledge is viable in certain scenarios. For example, we can compute pairwise similarity between images or texts. With suitable choices of similarity functions, we can form positive semidefinite kernel matrices and use them in DPP inference problems, such as extracting the most diverse subset as in eq.~(\ref{eDPPMAP}).  Another possible route is unsupervised learning of kernel matrices, again, leveraging insights about the intrinsic properties of the items. On the other end, it would be more effective to learn those parameters from data. 

Suppose we are given a training set $\{(\ground_n, \vy_n)\}$, where each ground set $\ground_n$  is annotated with its most diverse subset $\vy_n$. How can we discover the underlying parameters $\mL$ or $\mK$? Note that different ground sets need not have overlap. Thus, directly specifying kernel values for every pair of items is unlikely to be scalable. Instead, we will need to assume that either $\mL$ or $\mK$ for each ground set is represented by a shared set of parameters $\vtheta$. 

For items $i$ and $j$ in $\ground_n$, suppose their kernel values $K_{n_{ij}}$ can be computed as a function of $\vx_{n_{i}}$, $\vx_{n_{j}}$ and $\vtheta$, where $\vx_{n_{i}}$ and $\vx_{n_{j}}$ are features characterizing those items. %We will postpone a detailed discussion of this issue to section~\ref{sMKL} where we illustrate several different ways of parameterization.
   Our learning objective is to optimize $\vtheta$ such that $\vy_n$ is the most diverse subset in $\ground_n$, or attains the highest probability. This gives rise to the following maximum likelihood estimate (MLE)~\cite{kulesza2011learning},
\begin{equation}
\vtheta^{\textsc{mle}} = \argmax_{\vtheta} \sum_n \log P(\vy_n; \mL_n(\ground_n; \vtheta)),
\end{equation}
where $\mL_n(\ground_n; \vtheta)$ converts features in $\ground_n$ to the $\mL$ matrix for the ground set $\ground_n$.  MLE has been a standard approach for estimating DPP parameters. However, as we will discuss in section~\ref{sMargin}, it has important limitations.

%\BG{The MLE is a popular estimator for DPP~\cite{kulesza2011learning,kulesza2011k,affandi2012markov}. However, its performance is often limited by the number of training data, since in many set selection applications it is difficult to obtain well labeled subsets. Besides, MLE is inflexible in supporting customarily tailored error functions.}
 
Next, we introduce our method for learning the parameters. We first present our multiple kernel based representation of the $\mL$ matrix and then the large-margin based estimation.
%DPP is a stochastic process arising from random matrix theory and quantum physics~\cite{macchi1975coincidence,kulesza2012determinantal}. It is characterized by the determinant of a kernel function or matrix. Here we focus on a special family of DPPs, the discrete and finite L-ensembles. 

%Due to the ``repulsive'' nature in determinant, DPP is often used in set selection problems where the diversity characteristic is required. As a concrete example, consider the probability of any pair of items being selected, $P(i,j\in\bm{Y})=\det{K_{\{i,j\}}}=K_{ii}K_{ij}-K_{ij}^2$. One can hence understand $K_{ij}$ as the ``repulsive'' strength between items $i$ and $j$; it decreases the probability of selecting these two items together.

%Next, we briefly describe an application of the DPPs on extractive summarization of documents and/or videos, through which we introduce some notations that will be used to derive our large-margin training approach.

% !TEX root = main.tex

\section{Our Approach} 
\label{sApproach}

Our approach consists of two components that are developed in parallel, yet work in concert: (1) the use of multiple kernel functions to represent the DPP; (2) applying the principle of large margin separation to optimize the  parameters. The former reduces the number of parameters to learn and thus is especially advantageous when the number of training samples is limited. The latter strengthens the advantage by optimizing objective functions that closely track subset selection errors. \couldcut{Empirical studies, to be presented in section~\ref{sExp}, validate the benefits of each component and confirm that the best results are attained when the two are combined.}    %While our approach is generally applicable, we focus on two practical tasks and exemplify the application of our approach to those real-world problems.

\subsection{Multiple kernel representation of a DPP} 
\label{sMKL}

Learning the $\mL$ or $\mK$ matrix for a DPP is an instance of learning kernel functions, as those matrices are positive semidefinite matrices, interpretable as kernel functions being evaluated on the items in the ground set. Thus, our goal is essentially to learn the right kernel function to measure similarity.

However, for many applications, similarity is just one of the criteria for selecting items. For instance, in the previous example of image retrieval, the retrieved images not only need to be diverse (thus different) but also need to have strong relevance to the query term. Similarly,
in  document summarization, the selected sentences not only need to be succinct and not redundant, but also need to represent the contents of the document~\cite{lin2010multi}. %Analogously, in the task of  video summarization, we  select a few representative frames from the original video sequence~\cite{goldman2006schematic}. There are often two key properties desired from the extracted items (sentences or frames): representativeness of each individual frame and diversity in a collection of frames.

Kulesza and Taskar~\cite{kulesza2011learning} propose to balance these two potentially conflicting forces with a decomposable $\mL$ matrix:
\begin{equation}
L_{ij} = q_i q_j S_{ij}= q_i q_j \vphi_i\T \vphi_j, \quad q_i= q(\vx_i) = \exp(\vtheta\T\vx_i), \quad \forall\ i, j \in \ground, 
\label{eDPPLDecompose}
\end{equation}
where $q_i$ is referred to as the \emph{quality} factor, modeling how representative  or relevant the selected items are. It depends on item $i$'s feature vector $\vx_i$, which encodes $i$'s contextual information and its \emph{representativeness} of  other items. For example, in document summarization, possible features are the sentence lengths,  positions of the sentences in the text, or others. $S_{ij}$, on the other hand, measures how \emph{similar} two sentences are, computed from a different set of features, $\vphi_i$ and $\vphi_j$, such as bag-of-words descriptors that represent each item's individual characteristics.  %The choice to use two types of features reflects the intent to model criteria beyond basic similarity.

However, prior work~\cite{kulesza2011learning} does not investigate whether this specific definition of similarity could be made optimal and adapted to the data, thus limiting the modeling power of the DPP largely to infer the quality $q_i$. Our empirical studies show that this limitation can be severe, especially when the modeling choice is erroneous (cf.\ section~\ref{sSynthetic}).

In this paper, we retain the aspect of quality modeling but improve the modeling of similarity $S_{ij}$ in two ways. First,  we use nonlinear kernel functions such as the Gaussian RBF kernel to determine similarity. Secondly, and more importantly, we combine several base kernels: 
\begin{equation}
S_{ij} = \sum_k \alpha_k \exp\{-\twonorm{\vphi_i - \vphi_j}/\sigma_k^2\} + \beta \vphi_i\T\vphi_j,
\label{eDPPMKL}
\end{equation}
where $k$ indexes the base kernels and $\sigma_k$ is a scaling factor.  The combination coefficients are constrained such that $\sum_k \alpha_k + \beta = 1$. They are optimized on the annotated data, either via maximum likelihood estimation or via our novel parameter estimation technique, to be described next.

\subsection{Large-margin estimation of DPP} 
\label{sMargin}

%nor reduce the chance of making those errors

Maximum likelihood estimation does not closely track discriminative errors~\cite{ng,vapnik,jebara}. While improving the likelihood of the ground-truth subset $\vy_n$, MLE could also improve the likelihoods of other competing subsets.  Consequentially, a model learned with MLE could have modes that are very different subsets yet are very close to each other in their probability values.  Having highly confusable modes is especially problematic for DPP's NP-hard MAP inference --- the difference between such modes can fall within the approximation errors of approximate inference algorithms such that the true MAP cannot be easily extracted.  

\paragraph{Multiplicative large margin constraints} To address these deficiencies, our large-margin based approach aims to maintain or increase the margin between the correct subset and alternative, incorrect ones.  Specifically, we formulate the following large margin constraints
\begin{equation}
\log P(\vy_n; \mL_n) \ge \max_{\vy \subseteq \ground_n} \log \ell(\vy_n, \vy)\, P( \vy; \mL_n) =  \max_{\vy \subseteq \ground_n} \log \ell(\vy_n, \vy) + \log P( \vy; \mL_n),
\label{eLMDPPConstraints}
\end{equation}
where $\ell(\vy_n, \vy)$ is a loss function measuring the discrepancy between the correct subset and an alternative $\vy$. We assume $\ell(\vy_n, \vy_n)=0$.

Intuitively, the more different $\vy$ is from $\vy_n$, the larger the gap we want to maintain between the two probabilities.  This way, the incorrect one has less chance to be identified as the most diverse one. Note that while similar intuitions have been explored in multiway classification and structured prediction, the margin here is \emph{multiplicative} instead of additive --- this is by design, as it leads to a tractable optimization over the exponential number of constraints, as we will explain later.

\paragraph{Design of the loss function} A natural choice for the loss function is the Hamming distance between $\vy_n$ and $\vy$, counting the number of disagreements between two subsets:
\begin{equation}
\ell_H(\vy_n, \vy) = \sum_{i\in \vy}\mathbb{I}[i\notin \vy_n]+\sum_{i\notin \vy}\mathbb{I}[i\in \vy_n].
\label{eDPPHammingLoss}
\end{equation}
In this loss function, failing to select the right item costs the same as adding an unnecessary item.  In many tasks, however, this symmetry does not hold. For example, in summarizing a document, omitting a key sentence has more severe consequences than adding a (trivial) sentence.

To balance these two types of errors, we introduce the generalized Hamming loss function,
\begin{equation}
\ell_\omega(\vy_n, \vy) = \sum_{i\in \vy}\mathbb{I}[i\notin \vy_n]+\omega \sum_{i\notin \vy}\mathbb{I}[i\in \vy_n].
\label{eDPPGeneralizedHamming}
\end{equation}
When $\omega$ is greater than 1,  the learning biases towards higher \emph{recall} to select as many items in $\vy_n$ as possible.  When $\omega$ is significantly less than 1, the learning biases towards high \emph{precision} to avoid incorrect items  as much as possible. Our empirical studies demonstrate such flexibility and its advantages in two real-world summarization tasks.

\paragraph{Numerical optimization} To overcome the challenge of dealing with an exponential number of constraints in   eq.~(\ref{eLMDPPConstraints}), we reformulate it as a tractable optimization problem. We first upper-bound the hard-max operation  with Jensen's inequality (i.e., softmax):
\begin{equation}
\log P(\vy_n; \mL_n) \ge \log \sum_{\vy \subseteq \ground} e^{\log \ell_\omega(\vy_n, \vy)\, P( \vy; \mL_n)} =\mathsf{softmax}_{\vy \subseteq \ground_n} \log \ell_\omega(\vy_n, \vy) + \log P( \vy; \mL_n).
\end{equation}
With the loss function $\ell_\omega(\vy_n, \vy)$, the right-hand-side is computable in polynomial time,
\begin{equation} 
\mathsf{softmax}_{\vy \subseteq \ground_n} \log \ell_\omega(\vy_n, \vy) + \log P( \vy; \mL_n) 
 =  \log \left(\sum_{i\notin \vy_n} K_{n_{ii}}+\omega\sum_{i\in \vy_n}(1-K_{n_{ii}})\right),
\end{equation}
where $K_{n_{ii}}$ is the $i$-th element on the diagonal of $\mK_n$, the marginal kernel matrix corresponding to $\mL_n$.  The detailed derivation of this result is in the supplementary material. Note that $\mK_n$ can be computed efficiently from $\mL_n$ through the identity eq.~(\ref{eDPPKernel}).

The $\mathsf{softmax}$ can be seen as a summary of all undesirable subsets (the correct subset $\vy_n$ does not contribute
to the weighted sum as $\ell_\omega(\vy_n, \vy_n) =0$). Our optimization balances this term with the likelihood of the target with the hinge loss function $[z]_+= \max(0, z)$
\begin{equation}
\min\quad   \sum_n \left[ -\log P(\vy_n; \mL_n)  + \lambda \log \left( \sum_{i\notin \vy_n} K_{n_{ii}}+\omega\sum_{i\in \vy_n}(1-K_{n_{ii}})\right) \right]_+
\end{equation}
where $\lambda \ge 0$ is a  tradeoff coefficient, to be tuned on validation datasets. Note that this objective function subsumes maximum likelihood estimation where $\lambda = 0$.  We optimize the objective function with subgradient  descent. Details are in the supplementary material.

\section{Related work}
\label{sRelated}

The DPP arises from random matrix theory and quantum physics~\cite{macchi1975coincidence,kulesza2012determinantal}. In machine learning, researchers have proposed different variations to improve its modeling capacity. Kulesza and Taskar introduced k-DPP to restrict the sets to have a constant size $k$~\cite{kulesza2011k}. Affandi \etal\ proposed a Markov DPP which offers diversity at adjacent time stamps~\cite{affandi2012markov}. A structured DPP was presented in~\cite{kulesza2011structured} to model trees and graphs. The MAP inference of DPP is generally NP-hard~\cite{ko1995exact}. Gillenwater \etal\ developed an 1/4-approximation algorithm~\cite{gillenwater2012near}. In practice, greedy inference gives rise to decent results~\cite{kulesza2011learning} though it lacks theoretical guarantees. Another popular alternative is to resort to fast sampling algorithms \cite{kang2013fast,kulesza2012determinantal}. 

In spite of much research activity surrounding DPPs, there is very little work exploring how to effectively learn the model parameters. MLE is the most popular estimator.  Compared to MLE, our approach is more robust to the number of training data or mis-specified models, and offers greater flexibility by incorporating customizable error functions.  A recent Bayesian approach works with the posterior over the parameters~\cite{affandi2014icml}. In contrast to that work, we develop a large-margin training approach for DPPs and directly minimize the set selection errors.  The large-margin principle has been widely used in classification~\cite{vapnik} and structured prediction~\cite{taskar,tsochantaridis2004support,taskar2004m3,sha2006hmm}, but its application to DPP is original. In order to make it tractable for DPPs, we use multiplicative rather than additive margin constraints.

% !TEX root = main.tex
\section{Experiments}
\label{sExp}

We validate our large-margin approach to learn DPP parameters with extensive empirical studies on both synthetic data and two real-world summarization tasks with documents and videos.  While DPP also has applications beyond summarization, this is a particularly good testbed to illustrate diverse subset selection: a compact summary ought to include high quality items that, taken together, offer good coverage of the source content.  We report key results in this section, and provide more extensive results in the supplementary material. 

%\KG{Check: Minor note: doesn't intro make point about doing ``better'' with less training data, yet results not really showign this?  We don't demonstrate MLE needing more training data explicitly.} \FS{Boqing/Harry: Can you check results with fewer training samples? Maybe we should emphasize MLE never gets better --- because the data is noisy but LM can always exploit better?} \BG{On the synthetic data the performance of MLE indeed cannot get close to that of LM-DPP, even given 1000 training data. However, when the data is clean (no noise), both MLE and LM-DPP perform nearly perfectly.}

\vspace*{-0.15in}
\subsection{Synthetic dataset}
\label{sSynthetic}

\paragraph{Data} Our ground set has 10 items, $\ground = \{\vx_1,\vx_2,\cdots,\vx_{10}\}$. For each item, we sample a 5-dimensional feature vector from a spherical Gaussian: $\vx_i \sim \mathcal{N}(\vzero, \mI)$. To generate the $\mL$ matrix for the DPP, we follow the model in eq.~\eqref{eDPPLDecompose}; for the parameter vector $\vtheta$ we sample from a spherical Gaussian, $\vtheta \sim \mathcal{N}(\vzero, \mI)$, and for the similarity we simply let $\vphi_i = \vx_i$ and compute $S_{ij} = \vphi_i\T\vphi_j$.

We identify the most diverse subset $\vy^*$ (eq.~\eqref{eDPPMAP}) via exhaustive search of all subsets, which is possible given the small ground set.  The resulting $\vy^*$ has 5 items on average. We then add noise by randomly (with probability 0.1) adding or dropping an item to or from $\vy^*$. We repeat the process of sampling another pair of the ground set and its most diverse set. We do so 200 times and use 100 pairs for holdout and 100 for testing.  We repeat the process to yield training sets of various sizes.

\paragraph{Evaluation metrics}  We evaluate the quality of the selected subset $\vy^{\textsc{map}}$ against the ground-truth $\vy^*$ using the F-score, which is the harmonic mean of precision and recall:
\begin{equation}
\text{F-score}=\frac{2\,\text{Precision}\times\text{Recall}}{\text{Precision}+\text{Recall}},\quad \text{Precision} = \frac{|\vy^{\textsc{map}} \cap \vy^*|}{|\vy^{\textsc{map}}|}, \quad \text{Recall} = \frac{|\vy^{\textsc{map}} \cap \vy^*|}{|\vy^*|}.
\end{equation}
All three quantities are between 0 and 1, and higher values are better.

\paragraph{Learning and inference} We compare our large-margin approach using the Hamming loss (eq.~\eqref{eDPPHammingLoss}) to the standard MLE method for learning DPP parameters.\footnote{Adding a zero-mean Gaussian prior over $\vtheta$ while learning with MLE, as in \cite{kulesza2011learning}, did not yield improvement.} All hyperparameters are tuned by cross-validation.  After learning, we apply MAP inference to the testing ground sets. 
%%%Given the small size of the ground sets, we exhaustively search for the most diverse subsets.

\begin{figure}[t]
\centering
\subfigure[{\small Learning $\vtheta$ only, with $S_{ij}$ correctly specified}]{\label{fLearnQ}\includegraphics[width=0.32\textwidth]{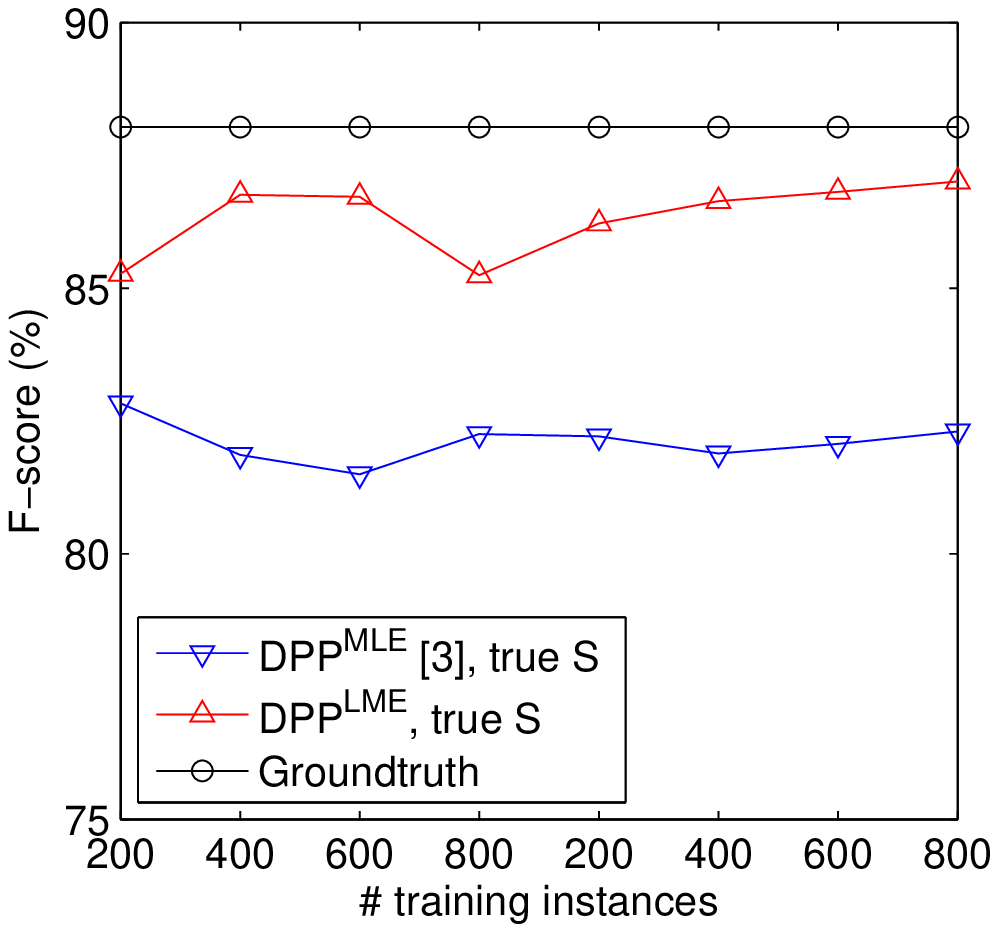}}
~
\subfigure[{\small Learning $\vtheta$ under mis-specified $S_{ij}$} (\# training instances = 200)]{\label{fLearnQWIncorrectS}\includegraphics[width=0.32\textwidth]{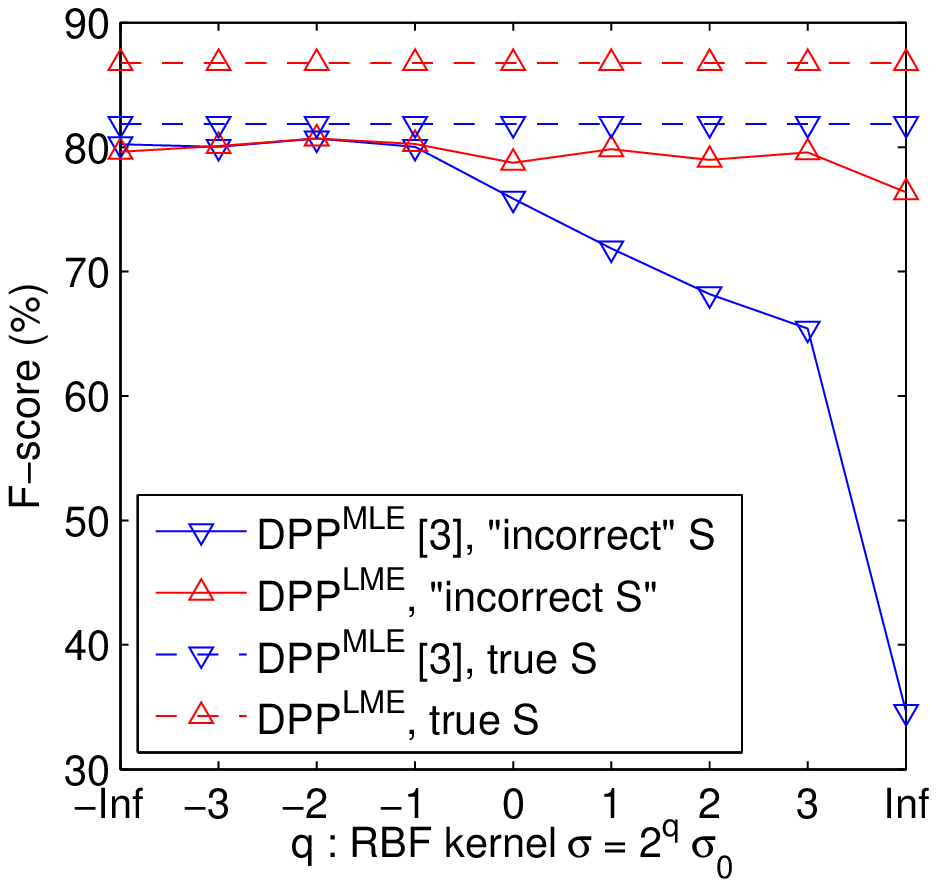}}
~
\subfigure[{\small Learning both $\vtheta$ and $S_{ij}$, with multiple kernel parameterization}]{\label{fLearnQSFscore}\includegraphics[width=0.32\textwidth]{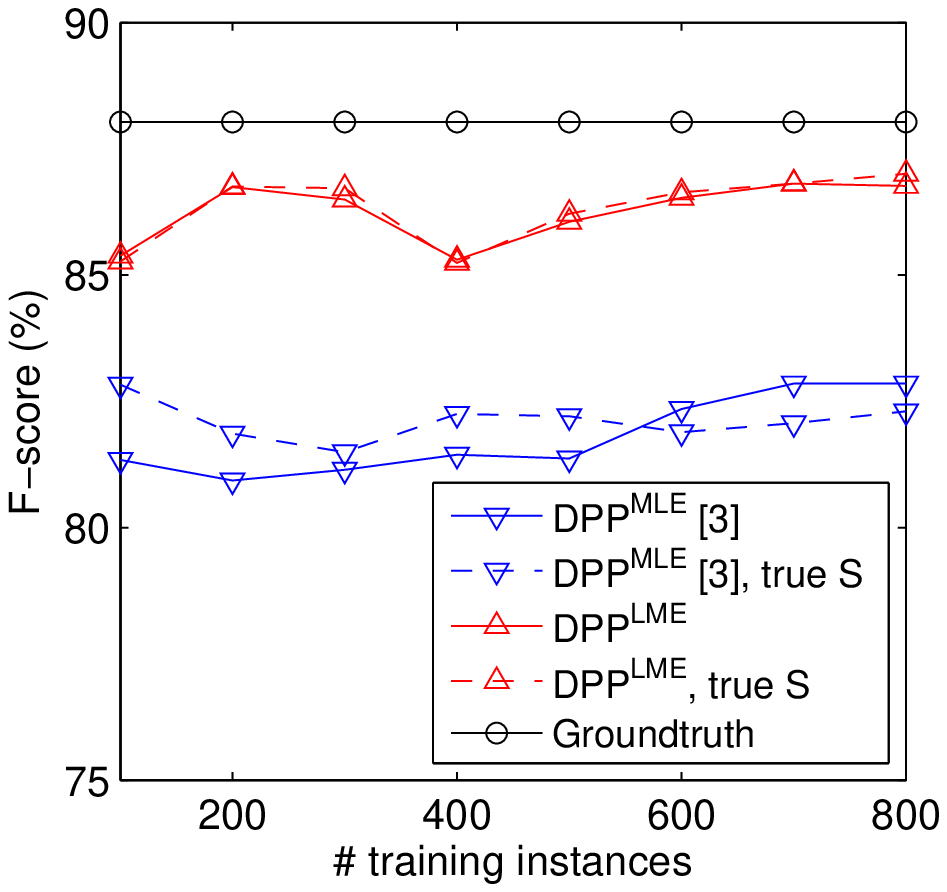}}\vspace*{-0.1in}
\caption{\small On synthetic datasets, our method $\textsc{dpp}^{\textsc{lme}}$ significantly outperforms the state-of-the-art parameter estimation technique $\textsc{dpp}^{\textsc{mle}}$~\cite{kulesza2011learning} in various learning settings.  See text for details. Best viewed in color.}\vspace*{-0.1in}
\label{fSynthetic}
\end{figure}

\paragraph{Results} The DPP is parameterized by two things: $\vtheta$ for the quality of the items, and $S_{ij}$ for the similarity among them. Since the ground-truth parameters are known to us, we conduct experiments to isolate the impact of learning either one.  %We do not measure whether or not the parameter values are recovered precisely since the MAP inference is scale-invariant with respect to $S_{ij}$.

Fig.~\ref{fLearnQ} contrasts the two methods when learning  $\vtheta$ only, assuming all $S_{ij}$ are known and the ground-truths are used. Our $\textsc{dpp}^{\textsc{lme}}$ method significantly outperforms $\textsc{dpp}^{\textsc{mle}}$. When the number of training samples is increased, the performance of our method generally improves and gets very close to the oracle's performance, for which the true values of \emph{both} $S_{ij}$ and $\vtheta$ are used.

Fig.~\ref{fLearnQWIncorrectS} examines the two methods in the setting of model mis-specification, where the $S_{ij}$ values deliberately deviate from the true values. Specifically, we set them to $\exp(-\|\vx_i-\vx_j\|^2_2/\sigma^2)$ where the bandwidth $\sigma$ varies from small to large, while the true values are $\vx_i\T\vx_j$. All methods generally suffer. However, our method is fairly robust to the mis-specification while $\textsc{dpp}^{\textsc{mle}}$ quickly deteriorates.  Our advantage is likely due to our method's focus on learning to reduce subset selection errors, whereas MLE focuses on learning the right probabilistic model (even if it is already mis-specified).

Fig.~\ref{fLearnQSFscore} compares the two methods when both $\vtheta$ and $S_{ij}$ need to be learned from the data. We apply our multiple kernel parameterization technique to model $S_{ij}$, as in eq.~\eqref{eDPPMKL}, except $\beta$ is set to be zero to avoid including the ground-truth. We see that our parameterization overcomes the problems of model mis-specification in Fig.~\ref{fLearnQWIncorrectS}, demonstrating its effectiveness in approximating unknown similarities. In fact, both learning methods match the performance of the corresponding methods with ground-truth similarity values, respectively.  Nonetheless, our large-margin estimation still outperforms MLE significantly.

In summary, our results on synthetic data are very encouraging.  Our multiple kernel parameterization avoids the pitfall of model mis-specification, and the large margin estimation outperforms MLE due to its ability to track selection errors more closely.

\subsection{Document summarization}

Next we apply  DPP to the task of extractive multi-document summarization~\cite{dang2005overview,kulesza2011learning,lin2010multi}.   In this task, the input is a document cluster consisting of several documents on a single topic. The desired output is a subset of the sentences in the cluster that serve as a summary for the entire cluster.  Naturally, we want the sentences in this subset to be both representative and diverse. 

\paragraph{Setup} We use the text data from Document Understanding Conference (DUC) 2003 and 2004~\cite{dang2005overview} as the training and testing sets, respectively. There are 60 document clusters in DUC 2003 and 50 in DUC 2004, each collected over a short time period on a single topic. A cluster includes 10 news articles and on average 250 sentences. Four human reference summaries are provided along with each cluster.  Following prior work, we generate the oracle/ground-truth summary by identifying a subset of the original sentences that best agree with the human reference summaries~\cite{kulesza2011learning}. On average, the oracle summary consists of 5 sentences.   As is standard practice, we use the oracles only during training.  During testing, the algorithm output is evaluated against each of the four human reference summaries separately, and we report the average accuracy~\cite{dang2005overview,kulesza2011learning,lin2010multi}. 

We use the widely-used evaluation package ROUGE~\cite{lin2004rouge}, which scores document summaries based on $n$-gram overlap statistics. We use  ROUGE 1.5.5 along with WordNet 2.0, and report the F-score (F), Precision (P), and Recall (R) of both unigram and bigram matchings, denoted by ROUGE-1X and ROUGE-2X respectively (X $\in$ \{F, P, R\}).  %Among them, ROUGE-1F is used in the MBR inference (eq.~(\ref{eMBR})). 
Additionally, we limit the maximum length of each summary to be 665 characters to be consistent with existing work~\cite{dang2005overview}.  This yields 5 sentences on average for subsets generated by our algorithm.

\begin{table}
\centering
\small
\caption{\small Accuracy on document summarization. Our methods outperform others with statistical significance. } \label{tDocSum}
{\small
\begin{tabular}{l|ccc|ccc}
Method & \textsc{rouge-1f} & \textsc{rouge-1p} & \textsc{rouge-1r} & \textsc{rouge-2f} & \textsc{rouge-2p} & \textsc{rouge-2r} \\
\hline
PEER 35~\cite{dang2005overview} & 37.54 & 37.69 & 37.45 & 8.37 & -- & -- \\
PEER 104~\cite{dang2005overview} & 37.12 & 36.79 & 37.48 & 8.49 & -- & -- \\
PEER 65~\cite{dang2005overview} & 37.87 & 37.58 & 38.20 & 9.13 & -- & -- \\
\textsc{dpp}$^{\textsc{mle}}$+\textsc{cos}~\cite{kulesza2011learning} &  37.89\scriptsize{$\pm$0.08} & 37.37\scriptsize{$\pm$0.08} & 38.46\scriptsize{$\pm$0.08} & 7.72\scriptsize{$\pm$0.06} & 7.63\scriptsize{$\pm$0.06} & 7.83\scriptsize{$\pm$0.06} \\
Ours (\textsc{dpp}$^{\textsc{lme}}$+\textsc{cos}) & 38.36\scriptsize{$\pm$0.09} & 37.72\scriptsize{$\pm$0.10} & 39.07\scriptsize{$\pm$0.08} & 8.20\scriptsize{$\pm$0.07} & 8.07\scriptsize{$\pm$0.07} & 8.35\scriptsize{$\pm$0.07} \\
Ours (\textsc{dpp}$^{\textsc{mle}}$+\textsc{mkr}) & 39.14\scriptsize{$\pm$0.08} & 39.03\scriptsize{$\pm$0.09} & 39.31\scriptsize{$\pm$0.09} & 9.25\scriptsize{$\pm$0.08} & 9.24\scriptsize{$\pm$0.08} & 9.27\scriptsize{$\pm$0.08} \\
Ours (\textsc{dpp}$^{\textsc{lme}}$+\textsc{mkr}) & \textbf{39.71}\scriptsize{$\pm$0.05} & \textbf{39.61}\scriptsize{$\pm$0.08} & \textbf{39.87}\scriptsize{$\pm$0.06} & \textbf{9.40}\scriptsize{$\pm$0.08} & \textbf{9.38}\scriptsize{$\pm$0.08} & \textbf{9.43}\scriptsize{$\pm$0.08}\\
\hline
\end{tabular}
} \vspace{-7pt}
\end{table}

To allow the fairest comparison to existing DPP work for this task, we use the same features designated in~\cite{kulesza2011learning}. To model quality, the features are the sentence length, position in the original document, mean cluster similarity, LexRank~\cite{erkan2004lexrank}, and personal pronouns. To model the similarity, the features are the standard normalized term frequency-inverse document frequency (tf-idf) vectors.

%%%, normalized to have unit length. 

%Further, we z-score them to have zero mean and one standard deviation within a document cluster. The traditional  is extracted for each sentence to serve as the similarity features ${\phi}$. We L2-normalize it so that $\|{\phi}\|_2^2=1$. 

\paragraph{Learning} We consider two ways of modeling similarities. The first one is to use the cosine similarity (\textsc{cos}) between feature vectors, as in~\cite{kulesza2011learning}.   The second is our multiple kernel based similarity (\textsc{mkr}, eq.~\eqref{eDPPMKL}). For \textsc{mkr}, the bandwidths are $\sigma=2^q$, $q=-6,-5,\cdots,6$, and the combination coefficients are learned on the data.  We implement the method in~\cite{kulesza2011learning} as a baseline (\textsc{dpp}$^{\textsc{mle}}$+\textsc{cos}).  We also test an enhanced variant of that method by replacing its cosine similarity with our multiple kernel based similarity (\textsc{dpp}$^{\textsc{mle}}$+\textsc{mkr}). 

\paragraph{Results}
Table~\ref{tDocSum} compares several DPP-based methods, as well as the top three results (PEER 35, 104, 65) from the DUC 2004 competition, which are not DPP-based (``-'' indicates results not available). Since the DPP MAP inference is NP-hard, we use a sampling technique to extract the most diverse subset~\cite{kulesza2012determinantal}. We run inference 10 times and report the mean accuracy and standard error. 

The state-of-the-art MLE-trained DPP model ($\textsc{dpp}^{\textsc{mle}}$+\textsc{cos})~\cite{kulesza2011learning} achieves about the same performance as the best PEER results of DUC 2004. We obtain a noticeable improvement by applying our large-margin estimation ($\textsc{dpp}^{\textsc{lme}}$+\textsc{cos}). By applying multiple kernels to model similarity, we obtain significant improvements (above the standard errors) for both parameter estimation techniques. In particular, our complete method, $\textsc{dpp}^{\textsc{lme}}$+\textsc{mkr}, attains the best performance across all the evaluation metrics.

\begin{table}
\centering
\caption{{\small Accuracy on video summarization.  Our method performs the best and allows precision-recall control.}} \label{tVideoSum}
\small
\begin{tabular}{c|c|c|c|c|c|c}
\hline
\multirow{2}{*}{Metric}& \multirow{2}{*}{VSUMM1~\cite{de2011vsumm}} & \multirow{2}{*}{VSUMM2~\cite{de2011vsumm}} & \multirow{2}{*}{\textsc{dpp}$^{\textsc{mle}}$+\textsc{mkr}} &\multicolumn{3}{|c}{Ours (\textsc{dpp}$^{\textsc{lme}}$+\textsc{mkr})}\\  \cline{5-7}
& & & &  $\omega=1/64$ & $\omega=1$ & $\omega=64$ \\ \hline
F-score & 70.25 & 68.20 & 72.94$\pm$0.08  & 71.25$\pm$0.09 & \textbf{73.46}$\pm$0.07 & 72.39$\pm$0.10 \\
Precision & 70.57 & 73.14 & 68.40$\pm$0.08 & \textbf{74.00}$\pm$0.09 & {69.68}$\pm$0.08 & {67.19}$\pm$0.11 \\
Recall & 75.77 & 69.14 & 82.51$\pm$0.11 & 72.71$\pm$0.11 & 81.39$\pm$0.09 & \textbf{83.24}$\pm$0.09\\ \hline
\eat{
Method & F-score & Precision & Recall\\
\hline
VSUMM1~\cite{de2011vsumm} & 70.25 & 70.57 & 75.77 \\
VSUMM2~\cite{de2011vsumm} & 68.20 & 73.14 & 69.14 \\
\textsc{dpp}$^{\textsc{mle}}$+\textsc{mkl} & 72.94$\pm$0.08 & 68.40$\pm$0.08 & 82.51$\pm$0.11 \\
Ours (\textsc{dpp}$^{\textsc{lme}}$+\textsc{mkl} w/ $\omega=1/64$) & 71.25$\pm$0.09 & \textbf{74.00}$\pm$0.09 & 72.71$\pm$0.11 \\
Ours (\textsc{dpp}$^{\textsc{lme}}$+\textsc{mkl} w/ $\omega=1$)  & \textbf{73.46}$\pm$0.07 & {69.68}$\pm$0.08 & {81.93}$\pm$0.09 \\
Ours (\textsc{dpp}$^{\textsc{lme}}$+\textsc{mkl} w/ $\omega=64$)  & 72.39$\pm$0.10 & 67.19$\pm$0.11 & \textbf{83.24}$\pm$0.09 \\
}
\end{tabular}
\vspace{-5pt}
\end{table}

\subsection{Video summarization}

Finally, we demonstrate the broad applicability of our  method by applying it to video summarization.  In this case, the goal is to select a set of representative and diverse frames from a video sequence.

%We randomly split the data into five folds, take one fold as the test set and the remaining as the training set. We test on every fold and report the averaged results.

\paragraph{Setup} The dataset consists of 50 videos from the Open Video Project (\textsc{ovp})\footnote{The Open Video Project: {www.open-video.org}}. They are 30fps, 352$\times$240 pixels, vary from 1 to 4 minutes, and are distributed across several genres including documentary, educational, historical, etc.  We use the provided ground truth key frame summaries~\cite{de2011vsumm}, where each video is labeled by five annotators independently. %There is no restriction on the number of frames to select. 
We perform 5-fold validation and report the average result.  We apply several preprocessing steps to remove frames that are trivially redundant (due to high temporal correlation) or of low visual quality.  We use a similar procedure as in the document summarization task to generate the ground-truth subsets. On average, the ground-truth has  9 frames (in contrast, our method yields subsets from 5 to 20 frames).  We use the public evaluation package VSUMM to evaluate the system-generated summary frames and again compute Precision, Recall and F-score~\cite{de2011vsumm}. 
%This package includes an automatic step of aligning two subsets first. 
  More details are in the supplementary material.

\paragraph{Features} We extract from each frame a color histogram and SIFT-based Fisher vector~\cite{lowe2004distinctive,perronnin2007fisher} to model pairwise frame similarity $S_{ij}$. The two features are combined via our multiple kernel representation. To model the quality of each frame, we extract both intra-frame and inter-frame representativeness features. They are computed on the  saliency maps~\cite{Rahtu2010sal,Hou2012PAMI} and include the mean, standard deviation, median, and quantiles of the maps as well as the the  visual similarities between a frame and its neighbors. We z-score them within each video sequence.

\paragraph{Results} Table~\ref{tVideoSum} compares several methods for selecting key frames: an unsupervised clustering method VSUMM~\cite{de2011vsumm} (we implemented its two variants, offering a degree of tradeoff between precision and recall, and finely tuned the parameters), $\textsc{dpp}^{\textsc{mle}}$ with a multiple kernel parameterization of $S_{ij}$, and our margin-based approach.  \couldcut{We do not report modeling with the cosine similarity as in the document summarization --- preliminary experiments indicate that choice does not perform well as the cosine similarity is especially designed for tf-itf types of feature vectors.}  For our method, we illustrate its flexibility to target different operating points, by varying the tradeoff constant $\omega$ in the generalized Hamming distance loss function eq.~\eqref{eDPPGeneralizedHamming}. Recall that higher values of $\omega$ will promote higher recall, while lower promote higher precision.

The results clearly demonstrate the advantage of our approach, particularly in how it offers finer control of the tradeoff between precision and recall.  By adjusting $\omega$, our method performs the best in each of the three metrics and outperforms the baselines by a statistically significant margin measured in the standard errors. Controlling the tradeoff is quite valuable in this application; for example, high precision may be preferable to a user summarizing a video he himself captured (he knows what appeared in the video, and wants a noise-free summary), whereas high recall may be preferable to a user summarizing a video taken by a third party (he has not seen the original video, and prefers some noise to dropped frames).   More detailed analysis, including exemplar video frames, are provided in the supplementary material.

% !TEX root = main.tex

\vspace*{-0.1in}
\section{Conclusion}
\label{sConclusion}

The determinantal point process (DPP) offers a powerful and probabilistically grounded approach for selecting diverse subsets. We proposed a novel technique for learning DPPs from annotated data.  In contrast to the status quo of maximum likelihood estimation, our method is more flexible in modeling pairwise similarity and avoids the pitfall of model mis-specification.  Empirical results demonstrate its advantages on both synthetic datasets and challenging real-world summarization applications.

%In this paper, we have proposed a large-margin training approach to DPP and demonstrated its effectiveness on both synthetic and challenging real applications (document summarization and video summarization). In contrast to MLE, our approach directly minimizes set selection errors. We have empirically shown that it is more robust than MLE to the number of training data or mis-specified models. Moreover, it also introduces more flexibilities to DPP. By using different error functions one can conveniently tailor the trained DPP models to have either high precision or high recall, depending on the application scenarios.

\small{
\bibliographystyle{unsrt}
\bibliography{main}
}

% Begin the merging of supplementary material

\section*{Appendix}
\appendix
%\begin{center}
%\toptitlebar
%\textbf{\LARGE Supplementary Material}
%\bottomtitlebar
%\end{center}
% !TEX root = supp.tex

%In this Supplementary Material, we provide extra details on the following:
%\begin{itemize}
% \item Sec.~\ref{aHamming}: deriving the softmax, eq.~(13) in the main text. We also show how to efficiently compute the marginal probability $P_{\{i\}}$ in the eq.~(13).
% \item Sec.~\ref{aSubgradient}: subgradients of the objective function of our large-margin DPP (cf.\ eq.~(14)).
% \item Sec.~\ref{aInference}: details of the minimum Bayes risk decoding, which is used as the inference algorithm in document and video summarization. 
% \item Sec.~\ref{aVIdeo}: extra details and results of video summarization. We provide the details of generating the oracle video summaries and how to evaluate summarization results against user summaries. We also present more quantitative and qualitative results on balancing precision and recall via our large-margin DPP.
%\end{itemize}

% !TEX root = supp.tex

\section{Calculating the softmax (cf.\ eq.~(13))}  \label{aHamming}
In the main text, we use softmax to deal with the exponential number of large-margin constraints and arrive at eq.~(13). Here we show how to calculate the right-hand side of eq.~(13).

Firstly, we compute $\sum_{\vy\subseteq\ground_n}\ell_\omega(\vy_n,\vy)P(\vy;\mL_n)$ as follows
\begin{align}
\sum_{\vy\subseteq\ground_n}\ell_\omega(\vy_n,\vy)P(\vy;\mL_n) &= \sum_{\vy\subseteq\ground_n}\left[\sum_{i:i\in \vy}\mathbb{I}(i\notin \vy_n)+\omega\sum_{i:i\notin \vy}\mathbb{I}(i\in \vy_n)\right]P(\vy;\mL_n) \\
&= \sum_{i=1}^{N}\left[\sum_{\vy:i\in \vy}\mathbb{I}(i\notin \vy_n)P(\vy;\mL_n)+\omega\sum_{\vy:i\notin \vy}\mathbb{I}(i\in \vy_n)P(\vy;\mL_n)\right]\\
&= \sum_{i=1}^{N}\left[\mathbb{I}(i\notin \vy_n)P_{n_{\{i\}}} + \omega \mathbb{I}(i\in \vy_n)\left(1-P_{n_{\{i\}}}\right)\right]\\
&= \sum_{i:i\notin \vy_n}P_{n_{\{i\}}} + \omega \sum_{i:i\in \vy_n}(1-P_{n_{\{i\}}}) \\
&=\sum_{i:i\notin \vy_n}K_{n_{ii}} + \omega \sum_{i:i\in \vy_n}(1-{K_{n_{ii}}}),
\end{align}
where $P_{n_{\{i\}}}=K_{n_{ii}}$ is the marginal probability of selecting item $i$. Now we are ready to see
\begin{align}
\mathsf{softmax}_{\vy \subseteq \ground_n} \log \ell_\omega(\vy_n, \vy) + \log P( \vy; \mL_n) &= \log \sum_{\vy\subseteq\ground_n}\ell_\omega(\vy_n,\vy)P(\vy;\mL_n) \\
&= \log\left(\sum_{i:i\notin \vy_n}K_{n_{ii}} + \omega \sum_{i:i\in \vy_n}(1-{K_{n_{ii}}})\right).
\end{align}

Moreover, recall that $\mK=\mL(\mL+\mI)^{-1}$. Eigen-decomposing $\mL=\sum_{m}\lambda_{m}\bm{v}_{m}\bm{v}_{m}^{T}$,
we have 
\begin{align}
\mK=\mL(\mL+\mI)^{-1}=\sum_{m}\frac{\lambda_{m}}{\lambda_{m}+1}\bm{v}_{m}\bm{v}_{m}^{T}, \quad\text{and thus,} \quad K_{ii}=\sum_{m}\frac{\lambda_{m}}{\lambda_{m}+1}{v}_{mi}^{2}.
\end{align}

\section{Subgradients of the objective function (cf.\ eq.~(14))} \label{aSubgradient}
Recall that our objective function in eq.~(14) actually consists of a likelihood term $\mathcal{L}(\cdot)$ and the other term of undesirable subsets. Denote them respectively by
\begin{align}
\mathcal{L}(\bm{\theta},\bm{\alpha}; \ground_n,\vy_n) &\triangleq \log P(\vy_n;\mL_n) = \log\det(\mL_{n_{\vy_n}}) - \log\det(\mL_n+\mat{I}), \\
\mathcal{A}(\bm{\theta},\bm{\alpha}; \ground_n,\vy_n)  &\triangleq\log\left(\sum_{i\notin \vy_n}K_{n_{ii}} + \omega \sum_{i\in \vy_n} (1-{K_{n_{ii}}}) \right).
\end{align}
For brevity, we drop the subscript $n$ of $\mL_n$ and $K_{n_{ii}}$ and change $\vy_n$ to $\vy_{\star}$ in what follows.

To compute the overall subgradients, it is sufficient to compute the gradients of the above two terms, $\mathcal{L}$ and $\mathcal{A}$. Denoting by $\bm{\Theta}=\{\bm{\theta},\bm{\alpha},\beta\}$, we have
\begin{align}
\frac{\partial \mathcal{L}}{\partial {\Theta_k}} 
= \sum_{i,j} \frac{\partial \mathcal{L}}{\partial L_{ij}} \frac{\partial L_{ij}}{\partial {\Theta_k}} 
= \bm{1}^T \left(\frac{\partial \mathcal{L}}{\partial \mL} \circ \frac{\partial \mL}{\partial {\Theta_k}} \right) \bm{1}, 
\quad \frac{\partial\mathcal{A}}{\partial \Theta_k} = \bm{1}^T \left(\frac{\partial\mathcal{A}}{\partial{\mL}}\circ \frac{\partial \mL}{\partial {\Theta_k}} \right) \bm{1},  \label{eGradient}
\end{align}
where $\circ$ stands for the element-wise product between two matrices of the same size. We use the chain rule to decompose $\frac{\partial \mL}{\partial {\Theta_k}}$ from the overall gradients on purpose. Therefore, if we change the way of parameterizing the DPP kernel $\mL$, we only need care about $\frac{\partial \mL}{\partial {\Theta_k}}$ when we compute the gradients for the new parameterization. 

\subsection{Gradients of the quality-diversity decomposition}
In terms of the quality-diversity decomposition (c.f.\ eq.~(7) and (8)), we have
\begin{align}
\frac{\partial \mL}{\partial {\alpha}_k} = (\bm{q}\bm{q}^T)\circ S^k, \quad
\frac{\partial L_{ij}}{\partial{\theta}_k}=L_{ij}({x}_{ik}+{x}_{jk}), \quad \text{or} \quad \frac{\partial \mL}{\partial{\theta}_k} = \mL\circ(\mat{X}\bm{e}_k\bm{1}^T+\bm{1}\bm{e}_k^T\mat{X}^T) 
\end{align}
where $\bm{q}$ is the vector concatenating the quality terms $q_i$, $\mat{X}$ is the design matrix concatenating $\bm{x}_i^T$ row by row, and $\bm{e}_k$ stands for the standard unit vector with 1 at the $k$-th entry and 0 elsewhere. 

\subsection{Gradients with respect to the DPP kernel}

In what follows we calculate $\frac{\partial\mathcal{L}}{\partial \mL}$ and $\frac{\partial\mathcal{A}}{\partial \mL}$ in eq.~(\ref{eGradient}). Noting that eq.~(\ref{eGradient}) sums over all the $(i,j)$ pairs, we therefore do not need bother taking special care of the symmetric structure in $\mL$.

We will need map $\mL_{\vy_\star}$ ``back'' to a matrix $\mat{M}$ which is the same size as the original matrix $\mL$, such that $\mat{M}_{\vy_\star}=\mL_{\vy_\star}$ and all the other entries of $\mat{M}$ are zeros. We denote by $\langle{\mL_{\vy_\star}}\rangle$ such mapping, \ie, $\langle{\mL_{\vy_\star}}\rangle=\mat{M}$. Now we are ready to see,
\begin{align}
\frac{\partial\mathcal{L}}{\partial \mL}
=\frac{\partial\log\det(\mat{L}_{\vy_\star})}{\partial \mL} - \frac{\partial\log\det(\mL+\mI)}{\partial \mL} 
= \langle (\mL_{\vy_\star})^{-1}\rangle - (\mL+\mI)^{-1}.
\end{align}

It is a little more involved to compute 
\begin{align}
\frac{\partial\mathcal{A}}{\partial \mL} = \frac{1}{\sum_{i\notin \vy_\star}K_{ii} + \omega\sum_{i\in \vy_\star} (1-{K_{ii}})}\left[\sum_{i\notin \vy_\star}\frac{\partial K_{ii}}{\partial \mL} - \omega\sum_{i\in \vy_\star} \frac{\partial{K_{ii}}}{\partial \mL}\right], \label{ePartialAwL}
\end{align}
which involves $\frac{\partial{K_{ii}}}{\partial \mL}$. 

In order to calculate $\frac{\partial{K_{ii}}}{\partial \mL}$, we start from the basic identity~\cite{beyer1991crc} of 
\begin{align} 
\frac{\partial \mat{A}^{-1}}{\partial t}=-\mat{A}^{-1}\frac{\partial \mat{A}}{\partial t}\mat{A}^{-1}, \label{ePartialInverse}
\end{align}
followed by $\frac{\partial \mat{A}^{-1}}{\partial A_{mn}}=-\mat{A}^{-1}\mat{J}^{mn}\mat{A}^{-1}$,
where $\mat{J}^{mn}$ is the same size as $\mat{A}$. The $(m,n)$-th entry of $\mat{J}^{mn}$ is 1 and all else are zeros.

Let $\mat{A}=(\mL+\mI)$. Noting that $\mK=\mL(\mL+\mI)^{-1}=\mI-(\mL+\mI)^{-1}=\mI-\mat{A}^{-1}$ and thus $K_{ii}=1-\left[\mat{A}^{-1}\right]_{ii}$, we have,
\begin{align}
\frac{\partial K_{ii}}{\partial L_{mn}} 
=-\frac{\partial\left[\mat{A}^{-1}\right]_{ii}}{\partial L_{mn}}  
=-\frac{\partial\left[\mat{A}^{-1}\right]_{ii}}{\partial A_{mn}} 
=\left[\mat{A}^{-1}\mat{J}^{mn}\mat{A}^{-1}\right]_{ii}
= [\mat{A}^{-1}]_{mi}[\mat{A}^{-1}]_{ni} \label{ePartialKii}.
\end{align}

We can also write eq.~(\ref{ePartialKii}) in the matrix form,
\begin{align}
\frac{\partial K_{ii}}{\partial \mL} = [\mat{A}^{-1}]_{\cdot i}[\mat{A}^{-1}]_{\cdot i}^{T}
= \mat{A}^{-1}\bm{e}_i\bm{e}_i^T\mat{A}^{-1}
= \mat{A}^{-1}\mat{J}^{ii}\mat{A}^{-1},
\end{align}
where $[\mat{A}^{-1}]_{\cdot i}$ is the $i$-th column of $\mat{A}^{-1}$.

Overall, we arrive at a concise form by writing out the right-hand-side of eq.~(\ref{ePartialAwL}) and merging some terms,
\begin{align}
\sum_{i\notin \vy_\star}\frac{\partial K_{ii}}{\partial \mL}-\omega\sum_{i\in \vy_\star}\frac{\partial K_{ii}}{\partial \mL}  
= \mat{A}^{-1}\mI_\omega(\overline{\vy_\star})\mat{A}^{-1}
= (\mL+\mI)^{-1}\mI_\omega(\overline{\vy_\star})(\mL+\mI)^{-1}
\end{align}
where $\mI_\omega({\overline{\vy_\star}})$ looks like an identity matrix except that its $(i, i)$-th entry is $-\omega$ for $i\in \vy_\star$.

%\section{Details of Synthetic Experiments} 
%
%\begin{figure}
%  \centering
%    \includegraphics[width=\textwidth]{fig/learning_Q}
%  \caption{Learning the qualities $q_i$ in DPP by MLE~\cite{kulesza2011learning} and our large-margin training approach (DISC).} \label{fSynQ}
%\end{figure}
%
%\begin{figure}
%  \centering
%    \includegraphics[width=\textwidth]{fig/learning_QS}
%  \caption{Learning both qualities $q_i$ and similarities $S_{ij}$ in DPP.} \label{fSynMKL}
%\end{figure}
%
%\BG{We have provided sufficient details in the main text? I think the precision and recall curves in Fig.~\ref{fSynQ} and \ref{fSynMKL} are the only ``supplementary material'' here. Probably we even do not need have this section here.}

\section{Minimum Bayes Risk decoding} \label{aInference}
We conduct the MAP inference of DPP by brute-forth search on the synthetic data, and turn to the so called minimum Bayes risk (MBR) decoding~\cite{goel2000minimum,kulesza2012determinantal} for larger ground sets on real data.

The MBR inference samples subsets $\mathcal{S}=\{\vy_1,\cdots,\vy_\cst{T}\}$ from the learned DPP and outputs the one $\hat{\vy}$ which achieves the highest consensus with the others, where the consensus can be measured by different evaluation metrics depending on applications. We use the F-score in our case. Particularly,
\begin{align}
\hat{\vy} \leftarrow \arg\max_{\vy_{t'}\in\mathcal{S}} \frac{1}{\cst{T}}\sum_{t=1}^\cst{T} \textsc{F-score}(\vy_{t'},\vy_t). \label{eMBR}
\end{align}
Note that the MBR inference has actually introduced some degrees of flexibility to DPP (and to other probabilistic models). It allows users to infer the desired output according to different evaluation metrics. As a result, the selected subset is not necessarily the ``true'' diverse subset, but is biased towards the users' specific interests.

\section{Video summarization} \label{aVIdeo}

We provide details on 1) how to generate oracle summaries as the supervised information to learn DPPs and 2) how to evaluate system-generated summaries against user summaries. We also present more results on balancing the precision and recall through our large-margin DPP.

\subsection{Oracle summary}
In the OVP dataset, each video comes along with five user summaries $\vy_1, \vy_2, \cdots,\vy_5$~\cite{de2011vsumm}. Similar to document summarization~\cite{kulesza2011learning}, we extract an ``oracle'' summary $\vy_\star$ from the five user summaries using a greedy algorithm. Initialize $\vy_\star=\emptyset$. From the frames not in $\vy_\star$, we pick out the one $i$ which contributes the most to the marginal gain, 
\begin{align}
\textsc{vsumm}(\vy_\star\cup\{i\},\{\vy_1, \cdots,\vy_5\})-\textsc{vsumm}(\vy_\star,\{\vy_1, \cdots,\vy_5\}),
\end{align} 
where \textsc{vsumm} is the package developed in~\cite{de2011vsumm} to evaluate video summarization results. We postpone to Section~\ref{sCUS} for describing the evaluation scheme of \textsc{vsumm}. Namely, we select the oracle frames greedily for each video and stop until the marginal gain becomes negative. We evaluate the oracle summaries against users' and find that they achieve high precision and recalls, 84.1\% and 87.7\% respectively, validating that the oracle summaries are able to serve as good supervised targets for training DPP models. 

The above procedure allows a ``user-independent'' definition of a good oracle summary for learning. Of course if the application goal were to generate user-specific summaries catering to a particular user's taste, one would instead simply apply our framework with $\vy_\star$ set to be that particular user's selection.

\subsection{ \textsc{vsumm}: evaluating video summarization results} \label{sCUS}

%\begin{table}[t]
%\centering
%\caption{The consensuses among different users and those between user labeled video summaries and the oracles we generated ($p$: precision, $r$: recall).} \label{tOracle}
%\begin{tabular}{|c|cccccr|}
%\hline
%\%           & \textsc{user}$_1$ & \textsc{user}$_2$ & \textsc{user}$_3$ & \textsc{user}$_4$ & \textsc{user}$_5$ & \textbf{\textsc{oracles}}\\
%\hline
%($p, r$)    &  (79.0,  75.8)        & (76.8, 81.1)  & (76.7,   82.4) & (80.1,   75.1)& (79.3,   77.5) & \textbf{(84.1,   87.7)} \\
%\hline
%\end{tabular}
%\end{table}

We evaluate video summarization results using the \textsc{vsumm} package~\cite{de2011vsumm}. Given two sets of summaries/frames, it searches for the maximum number of matched pairs of frames between them. Two images are viewed as a matched pair if their visual difference is below a certain threshold. \textsc{vsumm} uses normalized color histograms to compute such difference. Besides, each frame of one set can be matched to at most one frame of the other set, and vice versa. After the matching procedure, one can hence develop different evaluation metrics based on the number of matched pairs. In our experiments, we define F-score, precision, and recall (cf.\ eq.~(15) of the main text).

\subsection{More results on balancing precision and recall}
\begin{figure}
\centering
\includegraphics[width=0.8\textwidth]{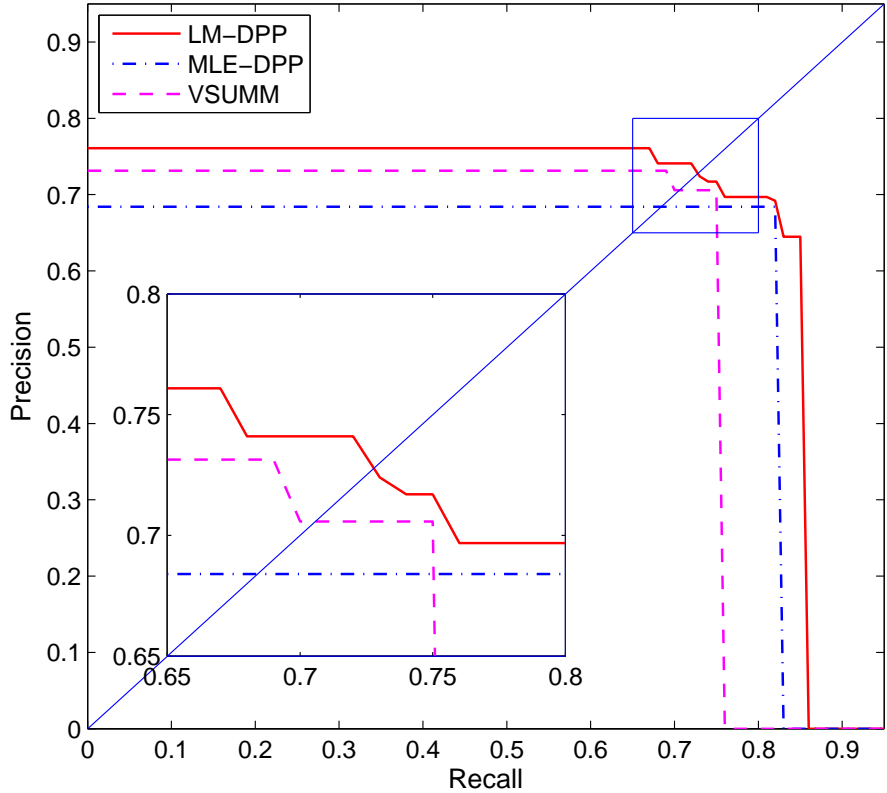}
\caption{Balancing precision and recall. Through our large-margin DPPs ($\textsc{dpp}^{\textsc{lme}}$),  we can balance precision and recall by varying $\omega$ in the generalized Hamming distance (cf.\ Section 3.2 in main text). In contrast, neither MLE nor VSUMM (the two variants in~\cite{de2011vsumm} are plotted together) is readily able to support such flexibility.} \label{fPR}
\end{figure}

\begin{figure}
\centering
\includegraphics[width=0.95\textwidth]{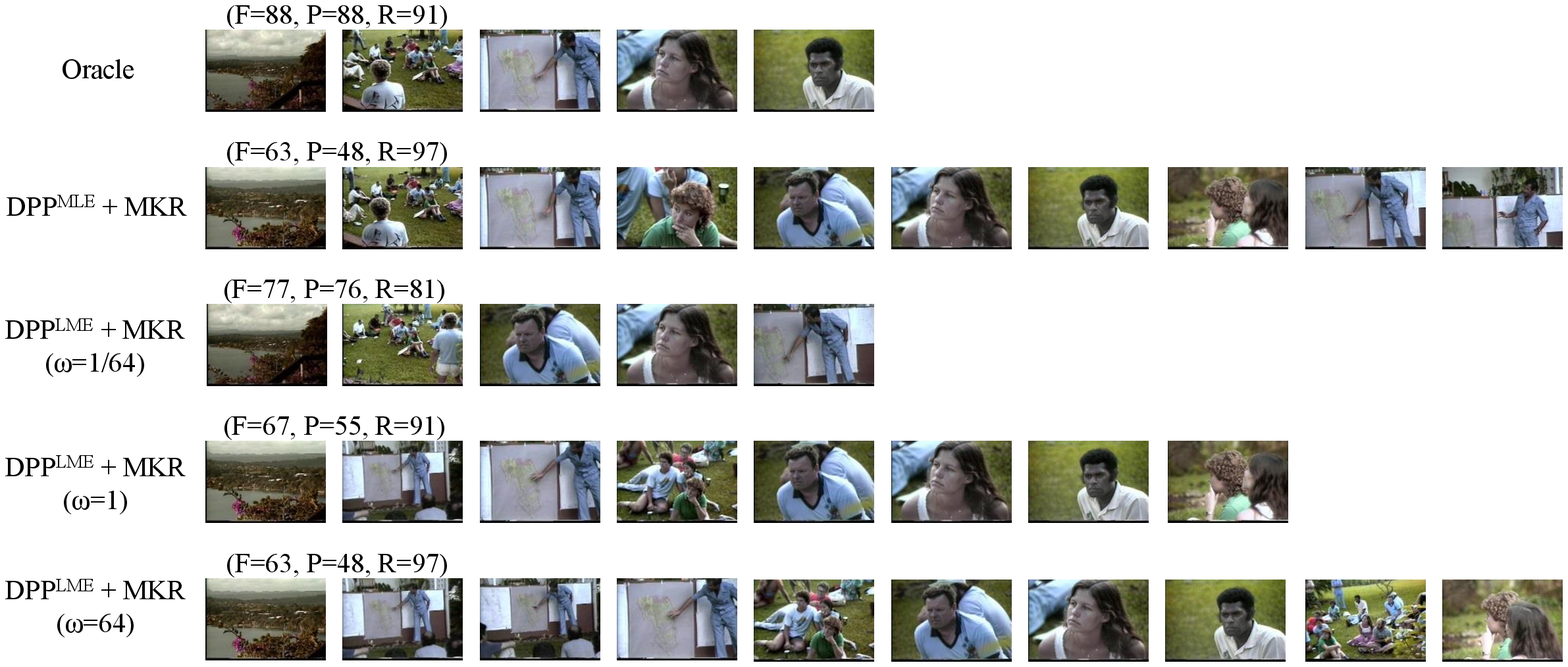}
\caption{Video summaries generated by $\textsc{dpp}^{\textsc{mle}}$ and our $\textsc{dpp}^{\textsc{lme}}$ with $\omega=1,\omega=2^6$, and $\omega=2^{-6}$, respectively. The oracle summary is also included for reference. } \label{fVideo61}
\end{figure}

We present more results here on balancing precision and recall through our large-margin trained DPPs ($\textsc{dpp}^{\textsc{lme}}$). By varying $\omega$ from $2^{-6}$ to $2^8$ in the generalized Hamming distance (cf.\ Section 3.2 in the main text), we obtain 8 pairs of (precision, recall) values. We apply uniform interpolation among them and draw the precision-recall curve in Fig.~\ref{fPR}. One can see that $\textsc{dpp}^{\textsc{lme}}$ is able to control the characteristics of the DPP generated summaries, baising them to either high precision or high recall and without sacrificing the other too much. Though MLE or VSUMM does not supply such modeling flexibility, we also include them in the figure for reference.

Besides, Fig.~\ref{fVideo61} shows some qualitative results. For this particular video, $\textsc{dpp}^{\textsc{mle}}$, $\textsc{dpp}^{\textsc{lme}}$ with $\omega=1$, and $\textsc{dpp}^{\textsc{lme}}$ with $\omega=2^6$ all give rise to high recalls. Their output summaries are pretty lengthy, and may be boring to some users who just want to grasp something interesting to watch. By turning down the weight to $\omega=2^{-6}$, our $\textsc{dpp}^{\textsc{lme}}$ dramatically improves the precision to 76\% (in contrast to the 48\% of $\textsc{dpp}^{\textsc{mle}}$).

\end{document}